\documentclass[10pt,twocolumn,letterpaper]{article}

\usepackage{cvpr}              % To produce the CAMERA-READY version
\definecolor{cvprblue}{rgb}{0.21,0.49,0.74}
\usepackage[pagebackref,breaklinks,colorlinks,allcolors=cvprblue]{hyperref}
\usepackage{algorithm}
\usepackage{algorithmic}
\usepackage{booktabs}
\usepackage{multirow}
\usepackage{xcolor}
\usepackage{newfloat}
\usepackage{listings}
\usepackage{bbm}
\usepackage[table]{xcolor}
\usepackage{amsmath,amsfonts,bm}

\def\eqref#1{equation~\ref{#1}}
\def\1{\bm{1}}

\def\vt{{\bm{t}}}

\def\mA{{\bm{A}}}

\def\mE{{\bm{E}}}

\def\mK{{\bm{K}}}

\def\mQ{{\bm{Q}}}

\def\mT{{\bm{T}}}

\def\mV{{\bm{V}}}

\def\mX{{\bm{X}}}
\def\mY{{\bm{Y}}}
\def\mZ{{\bm{Z}}}

\DeclareMathAlphabet{\mathsfit}{\encodingdefault}{\sfdefault}{m}{sl}
\SetMathAlphabet{\mathsfit}{bold}{\encodingdefault}{\sfdefault}{bx}{n}

\def\paperID{} % *** Enter the Paper ID here

\def\confName{CVPR}
\def\confYear{2026}

\title{Reasoning-Driven Anomaly Detection and Localization\\ with Image-Level Supervision}

\author{Yizhou Jin\textsuperscript{1,2} \hspace{0.5em} 
Yuezhu Feng\textsuperscript{1} \hspace{0.5em}
Jinjin Zhang\textsuperscript{1} \hspace{0.5em} 
Peng Wang\textsuperscript{1} \hspace{0.5em}
Qingjie Liu\textsuperscript{1,2}\footnotemark[1] \hspace{0.5em}
Yunhong Wang\textsuperscript{1,2}\footnotemark[1] \hspace{0.5em}
\\
\textsuperscript{1}State Key Laboratory of Virtual Reality Technology and Systems, Beihang University, Beijing, China \\
\textsuperscript{2}Hangzhou Innovation Institute, Beihang University, Hangzhou, China\\
{\tt\small \{yizhou.jin, fengyz, jinjin.zhang, pengwang717, qingjie.liu, yhwang\}@buaa.edu.cn}
}
\begin{document}
\maketitle
\footnotetext[1]{Joint corresponding authors.}

\begin{abstract}
Multimodal large language models (MLLMs) have recently demonstrated remarkable reasoning and perceptual abilities for anomaly detection.
However, most approaches remain confined to image-level anomaly detection and textual reasoning, while pixel-level localization still relies on external vision modules and dense annotations.
In this work, we activate the intrinsic reasoning potential of MLLMs to perform anomaly detection, pixel-level localization, and interpretable reasoning solely from image-level supervision, without any auxiliary components or pixel-wise labels.
Specifically, we propose \textbf{Re}asoning-Driven \textbf{A}nomaly \textbf{L}ocalization (ReAL), which extracts anomaly-related tokens from the autoregressive reasoning process and aggregates their attention responses to produce pixel-level anomaly maps.
We further introduce a \textbf{C}onsistency-\textbf{G}uided \textbf{R}easoning \textbf{O}ptimization (CGRO) module that leverages reinforcement learning to align reasoning tokens with visual attentions, resulting in more coherent reasoning and accurate anomaly localization.
Extensive experiments on four public benchmarks demonstrate that our method significantly improves anomaly detection, localization, and interpretability.
Remarkably, despite relying solely on image-level supervision, our approach achieves performance competitive with MLLM-based methods trained under dense pixel-level supervision. Code is available at https://github.com/YizhouJin313/ReADL.
\end{abstract}    
\section{Introduction}
Anomaly detection aims to identify and locate anomalous regions within images, which plays a critical role in industrial inspection. 
Traditional industrial anomaly detection methods~\cite{yao2024glad,guo2025dinomaly,liang2023omni,wu2025dfm,lei2023pyramidflow,liu2023simplenet,nafez2025patchguard,sun2025unseen,zavrtanik2021draem} are typically designed for single-product scenarios and require a large number of normal samples to train product-specific models. 
While effective under controlled conditions, these methods face limitations in real-world settings that involve strict data privacy constraints or frequent shifts in production lines~\cite{jin2024oner}.

Recent advances in multimodal large language models (MLLMs) have shown strong multimodal understanding and generation capabilities, opening new possibilities for industrial anomaly detection~\cite{li2024survey}.
MMAD~\cite{jiang2025mmad} established an evaluation benchmark for MLLMs in IAD, and Anomaly-OV~\cite{xu2025towards} further curated large-scale instruction-tuning datasets with chain-of-thought annotations, enabling MLLMs to generate human-aligned explanations for industrial defects. Building on these resources, several methods have adapted MLLMs to industrial anomaly detection. However, most existing approaches~\cite{li2025iad,li2025triad} remain confined to image-level anomaly detection and textual reasoning. Pixel-level localization, which is essential for actionable inspection, still relies on external vision modules.
For instance, AnomalyGPT~\cite{gu2024anomalygpt} employs a pretrained vision expert to generate anomaly maps. EIAD~\cite{zhang2025eiad} predicts anomaly bounding boxes and refines them with SAM~\cite{kirillov2023segment} to obtain segmentation masks. These staged designs depend on external, non-differentiable modules and often suffer from error propagation, misalignment between reasoning and localization, and increased deployment complexity.
To mitigate these issues, OmniAD~\cite{zhao2025omniad} introduces a text-as-mask encoding scheme that predicts segmentation masks via text tokens, providing a more unified end-to-end framework for anomaly detection, localization, and visualization.
Nevertheless, such end-to-end methods still rely on dense supervision, including pixel-level masks and high-quality reasoning annotations. Collecting this supervision is costly, sensitive to domain-specific bias, and can limit generalization to novel products and defect types.

In this work, we propose a reasoning-driven framework that unlocks the intrinsic anomaly understanding capacity of MLLMs. Our model performs joint anomaly detection, pixel-level localization, and interpretable reasoning under pure image-level supervision, without any auxiliary vision modules or pixel-level annotations. We design \textbf{Re}asoning-Driven \textbf{A}nomaly \textbf{L}ocalization (ReAL) to extract anomaly-relevant tokens from the autoregressive reasoning process and to construct pixel-level anomaly maps from their visual attentions. We further introduce \textbf{C}onsistency-\textbf{G}uided \textbf{R}easoning \textbf{O}ptimization (CGRO), which employs a reasoning–localization consistency reward to guide reinforcement learning and align reasoning tokens with their corresponding visual attentions. Together, these components enable a single MLLM to produce anomaly-aware reasoning and accurate localization using only image-level supervision.

Our contributions can be summarized as follows:
\begin{itemize}
\item We propose an end-to-end MLLM framework for anomaly understanding that unifies anomaly detection, pixel-level localization, and diagnostic reasoning, without relying on external vision modules or pixel-level annotations.
\item We introduce ReAL for reasoning-guided localization and CGRO for consistency-aware reinforcement learning. Together, they align textual reasoning with visual attention and enhance anomaly detection, localization, and interpretability under pure image-level supervision.
\item We show that, despite using only image-level supervision, our method achieves localization performance comparable to dense-supervision MLLM approaches on multiple benchmarks, demonstrating the effectiveness of the proposed framework.

\end{itemize}

\section{Related Work}

\subsection{Traditional Industrial Anomaly Detection}
Traditional industrial anomaly detection methods are predominantly unsupervised and can be grouped into three categories: reconstruction-based methods~\cite{yao2024glad,guo2025dinomaly,liang2023omni}, which detect anomalies via reconstruction errors; feature-embedding methods~\cite{wu2025dfm,lei2023pyramidflow,liu2023simplenet}, which measure deviations from normal feature distributions; and synthetic anomaly methods~\cite{nafez2025patchguard,sun2025unseen,zavrtanik2021draem}, which approximate decision boundaries by generating artificial defects. Despite their effectiveness, these approaches require large normal datasets and product-specific training, resulting in high computational and deployment costs.

To reduce data dependency, zero-shot and few-shot approaches leverage multimodal models for anomaly detection without extensive normal data. In zero-shot anomaly detection (ZSAD), most methods build upon CLIP~\cite{radford2021learning} to align image and text representations for anomaly scoring. For example, WinCLIP~\cite{jeong2023winclip} designs handcrafted prompt templates and extracts multi-level features for localization. Other CLIP-based methods improve textual guidance via prompt tuning~\cite{zhou2023anomalyclip,cao2024adaclip,ma2025aa} or representation refinement~\cite{gong2025fe,sadikaj2025multiads}. Beyond CLIP-based paradigms, LogSAD~\cite{Zhang_2025_CVPR} leverages MLLMs to generate matching proposals and model compositional reasoning for training-free anomaly detection.
Despite these advances, most methods still lack explicit modeling of interpretable cues such as anomaly categories, visual semantics, and causal factors~\cite{xu2025towards}, and remain limited by the reasoning capacity of their underlying models.
\subsection{MLLMs on Industrial Anomaly Detection}
Recent advances have enabled MLLMs~\cite{li2024survey} to incorporate task-specific modules for vision-centric tasks. However, existing benchmarks differ from industrial anomaly detection (IAD), which requires deeper product understanding and domain-specific reasoning. MMAD~\cite{jiang2025mmad} establishes a comprehensive benchmark for evaluating MLLMs in industrial inspection, while Anomaly-OV~\cite{xu2025towards} introduces the first visual instruction tuning dataset for ZSAD. IAD-R1~\cite{li2025iad} enhances inference through two-stage post-training with chain-of-thought reasoning, AD-FM~\cite{liao2025ad} reduces GRPO variance via multi-stage reasoning and refined rewards, and Triad~\cite{li2025triad} proposes a manufacturing-oriented IAD paradigm. Despite these efforts, MLLM-only approaches still struggle with precise pixel-level localization.

To integrate MLLMs into visual localization tasks, existing approaches typically leverage visual localization datasets during training and incorporate additional modules to extract localization information.
For instance, LISA\cite{lai2024lisa} employs a special [SEG] token to prompt a segmentation mask decoder, such as SAM~\cite{kirillov2023segment}, thereby improving reasoning-based segmentation performance.
The LVLM-based training-free visual grounding framework~\cite{kang2025your} directly leverages LVLM and achieves competitive segmentation performance.
Some studies have adapted MLLMs for anomaly segmentation and localization by using supplementary visual modules. AnomalyR1~\cite{chao2025anomalyr1} first predicts bounding box coordinates for anomaly detection and subsequently employs SAM to convert them into pixel-level masks. EIAD~\cite{zhang2025eiad} introduces a real-world abnormal multiple-choice QA dataset and performs segmentation via the [SEG] token, inspired by LISA.
For anomaly localization, AnomalyGPT~\cite{gu2024anomalygpt}, Myriad~\cite{li2023myriad}, and VELM~\cite{mokhtar2025detect} integrate auxiliary visual experts to provide fine-grained semantic information. These approaches introduce additional loss functions and architectural modifications, which increase the complexity of the processing pipeline and hinder scalability. 

Recently, end-to-end anomaly detection has been explored by OmniAD~\cite{zhao2025omniad}, which employs a hybrid decoder combining self-attention and deep convolution to learn normal patterns both locally and globally.
Nevertheless, such methods rely heavily on high-quality reasoning annotations and pixel-level masks, leading to substantial labeling costs and restricted generalization.
\begin{figure}[tp]
    \centering
    \includegraphics[width=0.99\linewidth]{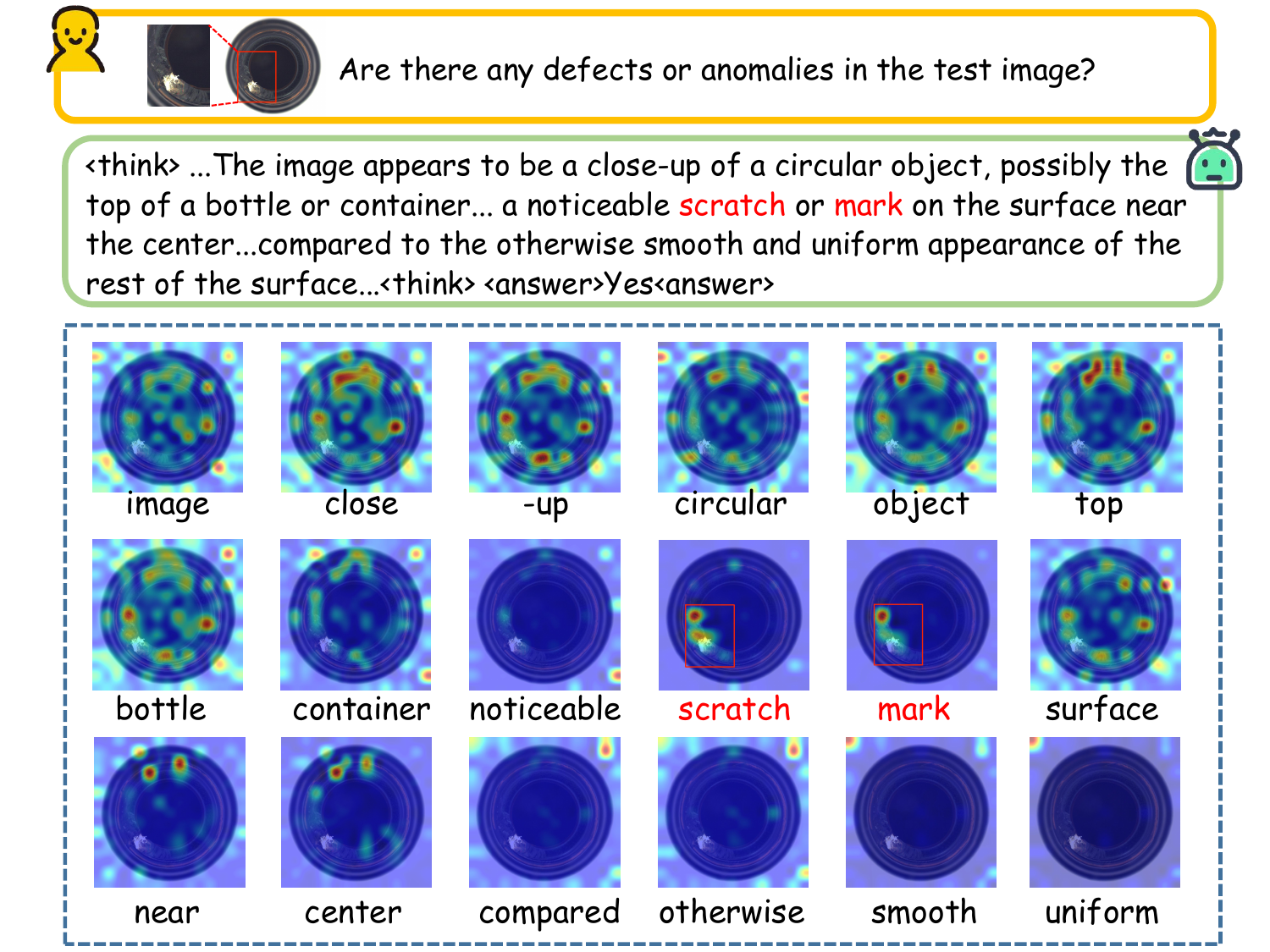}
    \caption{Visualization of token-to-patch attention maps for certain reasoning tokens. Only a small subset of tokens (highlighted in red) exhibits focused attention on the true anomaly regions (red boxes), and these tokens are also semantically related to anomaly concepts (e.g., “scratch”, “mark”).} 
    \label{fig:visual_attenton}
\end{figure}
\begin{figure*}[tp]
    \centering
    \includegraphics[width=0.95\linewidth]{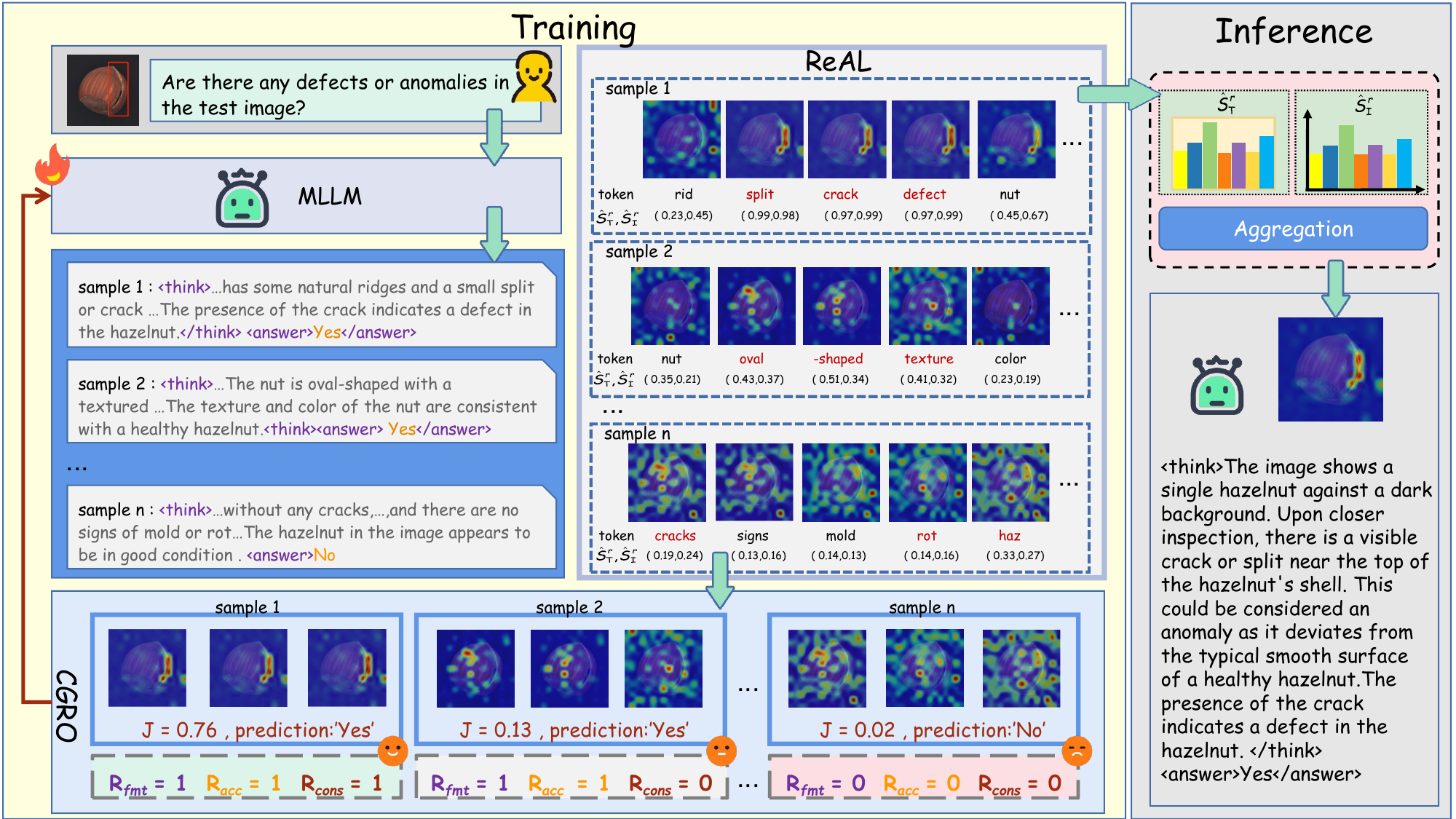}
    \caption{Overview of the proposed reasoning-driven anomaly detection framework. Given an input image, the MLLM generates a reasoning process and a final anomaly answer. The ReAL module selects anomaly-related tokens based on semantic relevance $S_\text{T}$ and spatial entropy $S_\text{I}$, then aggregates their visual attentions to obtain a pixel-level anomaly map. During training, these selected tokens are further used by the CGRO module to  leverage reinforcement learning jointly driven by a reasoning–localization consistency reward and R1-based accuracy and format rewards, enabling the model to align reasoning tokens with their corresponding visual attentions while simultaneously improving reasoning correctness, structural quality, and anomaly localization accuracy.} 
    \label{fig:framework}
\end{figure*}
\section{Anomaly Localization with MLLMs}
Recent studies have shown that the attention mechanism of MLLMs inherently captures how visual and textual information interact to generate output tokens~\cite{peng2023kosmos, ma2024groma}. 
In MLLM-based anomaly detection, given a test image and text prompts, the model typically produces a reasoning process that includes a global scene description, anomaly-related analysis, and a tentative conclusion, and then outputs the final answer. 
This raises a key question: when the model is generating the reasoning process, especially the tokens that describe anomalies, can we extract how the MLLM “focuses” on anomalous regions and leverage this focus for pixel-level anomaly localization?

Specifically, MLLMs take two inputs: an image $\mX_\text{v}$ and text prompts $\mX_\text{t}$. The text $\mX_\text{t}$ consists of a system prompt and a user question. The system prompt defines how the model should respond, guiding it to first generate a reasoning process and then provide a final answer. In this task, the user question remains fixed as: \texttt{Are there any defects or anomalies in the image?}
The image $\mX_v$ is divided into $P \times P$ patches and projected into visual embeddings $\mE_v$, while the text $\mX_t$ is tokenized into embeddings $\mE_t$. The concatenated embeddings $[\mE_v;\mE_t]$ are fed into the transformer-based MLLM, which autoregressively generates the output token sequence $\mT=\{ \vt_r\}_{r=1}^{n_\text{o}}$, where $n_\text{o}$ is the number of generated tokens. Decoding $\mT$ yields the output text $\mY$, which contains first the reasoning process and then the final answer.

Within each transformer block, multi-head self-attention models pairwise dependencies among all tokens. In layer $\ell$ and head $h$, the hidden states are projected into query $\mQ$, key $\mK$, and value $\mV \in \mathbb{R}^{(P\times P + n_\text{i}) \times d}$ matrices where $n_\text{i}$ is the number of input text tokens. The attention map is computed as:
\begin{equation}
\label{eq:prelim-attn}
    \text{Attn}^{\ell,h}(\mZ^{\ell-1}) = \text{softmax}\left(\frac{\mQ \mK^\top}{\sqrt{d_h}}\right),
\end{equation}
which measures the similarity between queries and keys.
For each decoding step, we aggregate across all heads and layers to obtain a global multimodal attention matrix $\mA\in \mathbb{R}^{N\times N}$, where $N=P\times P+n_\text{i}+n_\text{o}$. $\mA$ encodes interactions among $\mE_v$, $\mE_t$ and $\mT$. 

To investigate the impact of reasoning tokens, we visualize the token-to-patch attention map $\mA_{r,\text{I}}$ for each reasoning token $\vt_r$, computed as the average over all layers and heads. As shown in Fig.~\ref{fig:visual_attenton}, only a small fraction of reasoning tokens are capable of providing tangible and precise anomaly-aware spatial cues, and these tokens can in principle be used for anomaly localization. By inspection, we observe that such anomaly-relevant tokens are often associated with anomaly-related semantics (e.g., “scratch”, “mark”). However, without explicit supervision, identifying these informative tokens remains nontrivial, since the majority of reasoning tokens attend diffusely or focus on irrelevant regions, which in turn dilutes localization accuracy.
This observation motivates two further questions. First, can we automatically discover anomaly-relevant reasoning tokens and harness their attention patterns for pixel-level anomaly localization? Second, since the reasoning process already shows latent spatial grounding ability, can we deliberately activate this intrinsic reasoning potential to jointly improve anomaly detection, localization, and interpretability under limited supervision?

\section{Method}
% 待优化内容：attention的符号，两个score的名字，RL引入的逻辑。
To answer these questions, we propose a reasoning-driven framework that leverages the intrinsic perception and reasoning abilities of MLLMs to achieve end-to-end anomaly detection and localization under limited supervision, as illustrated in Fig.~\ref{fig:framework}. Specifically, we propose Reasoning-Driven Anomaly Localization (ReAL), which extracts anomaly-related tokens from the autoregressive reasoning process and aggregates their attention responses to produce pixel-level anomaly maps.
We further introduce a Consistency-Guided Reasoning Optimization (CGRO) module that leverages reinforcement learning driven by a reasoning–localization consistency reward to align reasoning tokens with visual attentions, resulting in more coherent reasoning and accurate anomaly localization.

\subsection{Reasoning-Driven Anomaly Localization}

\noindent\textbf{Identifying anomaly-relevant reasoning tokens.}
To distinguish reasoning tokens most relevant to anomaly localization, we evaluate each token from two complementary perspectives: 

\noindent\textit{Inter-modality semantic relevance.} This criterion measures how closely each reasoning token $\vt_r$ is related to the concept of anomaly. We compute the semantic relevance score $S_\text{T}^r$ by evaluating the attention of $\vt_r$ towards anomaly-related tokens in the input text (e.g., defect, anomaly or abnormal). The score is computed as: 
\begin{equation}
\label{eq:crit1}
    S_\text{T}^r = \sum_{e\in \mathcal{E}_\text{a}}\mA_{r,e},
\end{equation}
where $\mathcal{E}_\text{a}$ denotes the set of anomaly-related tokens in the textual input, and $\mathbf{A}_{r,e}$ represents the attention weight from reasoning token $\mathbf{t}_r$ to token $e$. A higher $S_\text{T}^r$ indicates stronger semantic relevance to anomalous concepts.

\noindent\textit{Intra-modality attention concentration.} To assess how focused a token's attention is on specific image regions, we compute the spatial entropy score $S_\text{I}^r$, which quantifies the concentration of attention across the image. A more concentrated attention implies the token is focusing on distinct, localized regions (such as anomalies), while a dispersed attention indicates a broader focus. To compute $S_\text{I}^r$, we first binarize the visual attention map $\mA_{r,\text{I}}$ by assigning a value of 1 to attention values above the threshold and 0 to those below. Next, we identify connected components $C$, where each component consists of neighboring pixels connected via 8-neighbors. The score is then calculated as:
    \begin{equation}
\label{eq:crit2}
    S_\text{I}^r = -\sum_{i=1}^{N} P(C_i) \log{P(C_i)},
\end{equation}
where $P(C_i)$ is the proportion of the total attention mass in component $C_i$. Lower values of $S_\text{I}^r$ indicate more concentrated attention, suggesting that the token is focusing on specific regions of the image, such as anomalies.
% \end{itemize}

By jointly considering these two criteria, we can reliably locate reasoning tokens that are both semantically associated with anomalies and visually concentrated on distinct regions, enabling pixel-level anomaly localization.

\noindent\textbf{Aggregating the reasoning-driven anomaly map.}
For each token $r$, we compute its semantic relevance $S_\text{T}^r$ and the visual concentration score $S_\text{I}^r$, normalize them to $\hat{S}_\text{T}^r$ and $\hat{S}_\text{I}^r$ (with inverse normalization for $S_\text{I}^r$, so a higher $\hat{S}_\text{I}^r$ indicates lower entropy). Tokens are retained if they satisfy $\hat{S}_\text{T}^r>\tau_\text{t}$ and $\hat{S}_\text{I}^r>\tau_\text{i}$, where the latter condition selects tokens with low spatial entropy and high semantic relevance to anomaly concepts. The composite token weight $w_r$ is defined as:
\begin{equation}
\label{eq:weight}
w_r = \mathbbm{1}_{\{\hat{S}_\text{T}^r>\tau_\text{t},\ \hat{S}_\text{I}^r>\tau_\text{i}\}} \left[ \alpha\hat{S}_\text{T}^r + \beta\hat{S}_\text{I}^r \right],
\end{equation}
where $\alpha$ and $\beta$ balance the contributions of semantic and visual information.

Finally, we aggregate the visual attention maps $\mA_{r,\text{I}}$ of the remaining filtered reasoning tokens, weighted by their corresponding $w_r$, to generate the reasoning-driven anomaly map:
\begin{equation}
\label{eq:rdam}
\mA_\text{RDAM} =
\sum_{r} w_r \cdot \mA_{r,\text{I}}.
\end{equation}
This aggregated map highlights the regions that receive the most attention from the most relevant reasoning tokens, allowing end-to-end anomaly localization.

\subsection{Consistency-Guided Reasoning Optimization}
With ReAL, the model can already produce pixel-level anomaly maps from reasoning tokens, which makes end-to-end anomaly localization possible without additional vision modules. However, most MLLMs cannot accurately locate and describe anomalies in the image without explicit human involvement \cite{xu2025towards}. 
In practice, many MLLM-based anomaly methods enhance performance by using large-scale instruction-tuning datasets tailored for anomaly detection, and pixel-level localization is often improved further through training with dense pixel-level annotations. 
% While effective, such supervision can be costly to obtain and may introduce domain-specific biases, which may limit generalization to novel products and defect types.
While effective, such supervision is costly and may introduce domain bias, limiting generalization to novel products and defect types.
Therefore, we adopt a reinforcement learning framework that relies only on image-level supervision, which is relatively easy to collect in industrial settings. Instead of relying on dense pixel-level annotations, the model learns by reflecting on its own predictions, gradually improving its anomaly perception and its ability to attend to the correct regions.
Under limited supervision, however, the reasoning generation of MLLMs lacks explicit constraints and can become inconsistent. For example, the model may correctly answer that an anomaly exists, while its reasoning still describes the image as normal. Such mismatches lead to unreliable interpretations and weaken the effectiveness of ReAL-based localization.

To address this issue, we propose Consistency-Guided Reasoning Optimization~(CGRO), which leverages reinforcement learning driven by a reasoning–localization consistency reward to align reasoning tokens with visual attentions, resulting in more coherent reasoning and accurate anomaly localization.

\noindent\textbf{Consistency reward.}
Based on the obtained weights $w$, we can identify the reasoning tokens most relevant to anomaly localization. The core idea is that, for anomaly images, these tokens should consistently focus on the regions corresponding to the anomalies, with visual attention concentrated in those areas. In contrast, for normal images, the attention should be more dispersed across the image, avoiding a focus on any specific region.

Concretely, we first select the top-$t$ reasoning tokens ranked by composite weight $w$.
For each selected token $r$, the 95th percentile of its visual attention map $\mathbf{A}_{r,\mathrm{I}}$ is used as the threshold;
spatial positions above this value are set to 1 and the rest to 0, yielding a binary support region $\Omega_r$.
The spatial consensus among these tokens is then measured by the Jaccard Index:
\begin{equation}
\label{eq:jaccard}
\mathcal{J} =
\frac{\big|\bigcap_{r=1}^{t} \Omega_{r}\big|}
{\big|\bigcup_{r=1}^{t} \Omega_{r}\big|},
\end{equation}
where $\mathcal{J}\in[0,1]$: larger values indicate that these tokens focus on highly overlapping regions, while smaller values indicate more dispersed or inconsistent attention. To encourage different behaviors for anomalous and normal images, we design a class-conditional consistency reward. For anomaly images $y=1$, we encourage high spatial consensus to promote coherent localization of defects; for normal images $y=0$, we encourage low spatial consensus to discourage spurious concentration on benign regions. This is implemented with a single indicator-based reward:
\begin{equation}
\label{eq:crl_reward}
R_\text{cons}= y\cdot  \mathbbm{1}(\mathcal(J)>\delta_{1})+(1-y)\cdot \mathbbm{1}(\mathcal(J)<\delta_{2}),
\end{equation}
where $\delta_{1}$ and $\delta_2$ are thresholds for anomaly and normal images, respectively, and $\mathbbm{1}{(\cdot)}$ denotes the indicator function.

\noindent\textbf{Training procedure.}
The model is optimized under the total reward defined as:
\begin{equation}
\label{eq:total_reward}
\mathcal{R}_\text{total}=\mathcal{R}_{\text{fmt}}+\mathcal{R}_{\text{acc}}+\mathcal{R}_{\text{cons}}.
\end{equation}
where $\mathcal{R}_{\text{fmt}}$ ensures well-structured outputs (i.e., reasoning followed by an answer), $\mathcal{R}_{\text{acc}}$ promotes correct image-level anomaly predictions, and $\mathcal{R}_{\text{cons}}$ enforces consistency between reasoning and visual evidence.

Following the group-based policy optimization~\cite{guo2025deepseek}, given a question–answer pair $(q,a)$, we sample a group of $G$ responses $\{o_i\}_{i=1}^{G}$ from the previous policy $\pi_{\theta_{\text{old}}}$. 
We update the policy by maximizing a clipped surrogate with a KL penalty:
\begin{equation}
\small
\label{eq:policy_obj}
\begin{aligned}
\mathcal{J}(\theta)
&= \mathbb{E}_{(q,a)\sim D,\,\{o_i\}\sim\pi_{\theta_{\text{old}}}}
\Bigg[
\frac{1}{G}\sum_{i=1}^{G}
\Big(
\min\!\Big(
\frac{\pi_{\theta}(o_i\mid q)}{\pi_{\theta_{\text{old}}}(o_i\mid q)}{a}_i,\;
\\
&\operatorname{clip}\!\Big(
\tfrac{\pi_{\theta}(o_i\mid q)}{\pi_{\theta_{\text{old}}}(o_i\mid q)},\,1-\varepsilon,\,1+\varepsilon
\Big){a}_i
\Big)
\;-\;\beta\,\mathbb{D}_{\mathrm{KL}}(\pi_{\theta}\Vert\pi_{\mathrm{ref}})
\Big)
\Bigg],
\end{aligned}
\end{equation}
where ${a}_i$ denotes the normalized advantage of each sample:
\begin{equation}
\label{eq:advantage}
{a}_i=\frac{r_i-\operatorname{mean}(\{r_1,\dots,r_G\})}{\operatorname{std}(\{r_1,\dots,r_G\})}.
\end{equation}
Through this optimization, the model learns to produce outputs that are well-formatted, accurate, and visually consistent.
The integration of the consistency reward enables the model to regulate its own reasoning process, resulting in reasoning that is semantically coherent, visually grounded, and more reliable for anomaly detection and localization.

% 增加Anomaly-R1 Anomaly-OV和IAD-R1 (Qwen3b)后的表格
\begin{table*}[htbp]
\centering
\tiny 
\caption{Comparison of image-level and pixel-level detection performance and reasoning performance under zero-shot settings. Detection results are reported as (AUROC, ACC), and reasoning results as (ROUGE-L, SBERT). The column `Anno.' denotes the supervision type used in training, where `I' represents image-level supervision, `P' denotes pixel-level annotations, and `T' indicates task-specific instruction-tuning data designed for anomaly detection. * indicates anomaly maps are generated with ReAL.}
\begin{tabular}{lccccccccccccc}
\toprule
 \multirow{2}{*}{\textbf{Model}} & \multirow{2}{*}{\textbf{Param.}}  & \multirow{2}{*}{\textbf{Anno.}} & \multicolumn{5}{c}{\textbf{Image-level}} & \multicolumn{5}{c}{\textbf{Pixel-level}}  & \multicolumn{1}{c}{\textbf{Reasoning}} \\
\cmidrule(lr){4-8} \cmidrule(lr){9-13} \cmidrule(lr){14-14}
& && \underline{\textbf{AVG}}&\textbf{SDD} & \textbf{DTD} & \textbf{WFDD} & \textbf{MVTec} & \underline{\textbf{AVG}} &\textbf{SDD} & \textbf{DTD} & \textbf{WFDD} & \textbf{MVTec}  & \textbf{MVTec-COT}\\
\midrule
% WinCLIP\\
% AdaCLIP\\
% AnomalyCLIP\\
% \midrule
GPT-4o-mini~\cite{hurst2024gpt} & / & / & 70.9,~71.4  & 66.2,~92.8 & 79.4,~70.3 & 66.4,~63.5& 71.5,~59.0& N/A & N/A & N/A& N/A & N/A  & 19.0,~65.2 \\
GPT-4o~\cite{hurst2024gpt} & / & / & 79.4,~85.3 & 	77.5,~90.7 &	89.2,~90.2 & 78.2,~ 78.6 & 	72.7,~81.7 & N/A & N/A & N/A& N/A & N/A & 19.9,~60.4 \\
%  GPT-4.1-nano & / & / & N/A & N/A & N/A& N/A & N/A &  &  \\
% GPT-4.1-mini & / & / & & &  &  &  & N/A & N/A & N/A&N/A &  \\
GPT-4.1 & / & / & \textbf{87.2},~\textbf{88.4} & 84.4,~92.1 &	94.3,~94.3 & 86.8,~86.0 & 83.3,~81.2& N/A & N/A & N/A & N/A & N/A & 20.8,~69.9 \\
% Claude-Sonnet-4& / & / & & &  &  &  & N/A & N/A & N/A&N/A &  \\
GPT-5-nano & / & / & 80.3,~81.6 &	82.6,~88.6 &	90.4,~87.2 & 80.2,~79.7 & 68.0,~70.7 & N/A & N/A & N/A & N/A & N/A &	14.0,~50.0 \\
Gemini-2.5-Flash~\cite{comanici2025gemini} & / & / & 79.9,~76.3 & 71.8,~49.5& 88.6,~91.3& 82.6,~82.2 & 76.5,~82.3 & N/A & N/A & N/A & N/A & N/A & 17.6,~68.1\\ 

\midrule
% LLaVA-OneVision-SI~\cite{li2024llava} & 0.5B & / & 53.8,~54.3 & 50.0,~95.5 &	54.3,~33.9 &52.7,~48.7 & 58.0,~39.1 & N/A & N/A & N/A & N/A & N/A & 13.1,~40.2 \\ 
LLaVA-OneVision-SI~\cite{li2024llava} & 7B & / & 78.3,~75.7 &	65.5,~67.8 &	91.5,~89.9 &	77.0,~75.0& 79.4,~70.1& N/A & N/A & N/A & N/A & N/A & 10.5,~39.1 \\
LLaVA-OneVision-OV~\cite{li2024llava} & 7B & / & 78.5,~81.4 & 	68.1,~94.7 &92.4,~89.1&75.5,~73.4& 77.8,~68.5 & N/A & N/A & N/A & N/A & N/A & 23.7,~70.5 \\
% LLaVa-1.5~\cite{liu2024improved} & 7B & / & N/A & N/A & N/A & N/A & N/A & 59.4,~60.2&55.3,~95.5 &	55.1,~35.5 &66.0,~63.1 &61.3,~46.7& / \\
LLaVA-1.6~\cite{liu2024llava} & 7B & / & 62.9,~63.5 & 67.0,~50.4 & 	70.2,~74.2 & 59.5,~60.6 & 54.8,~68.9 & N/A & N/A & N/A & N/A & N/A  & 18.6,~54.2 \\ 

Qwen2.5-VL~\cite{bai2025qwen2} & 3B & / & 52.9,~56.6 & 46.3,~69.9 & 	64.1,~51.3 & 50.0,~52.3 & 51.0,~52.8& N/A & N/A & N/A & N/A & N/A & 16.8,~49.0 \\
Qwen2.5-VL~\cite{bai2025qwen2} & 7B & / &63.4,~63.3&65.0,~93.8&	68.3,~56.2 &60.5,~57.2 &59.6,~45.8& N/A & N/A & N/A & N/A  & N/A & 20.8,~65.5 \\ 
Qwen2.5-VL~\cite{bai2025qwen2} & 32B & / &74.6,~71.9 & 82.4,~90.1 & 81.0,~73.0 &68.4,~70.5 &66.5,~54.0 & N/A & N/A & N/A & N/A & N/A & 20.0,~66.1 \\
Qwen2.5-VL~\cite{bai2025qwen2} & 72B & / & 81.1,~80.4& 84.5,~95.6 & 91.2,~87.6 & 74.5,~72.8&  74.1,~65.5 & N/A & N/A & N/A & N/A & N/A & 21.1,~67.5 \\ 
Qwen3VL-thinking~\cite{team2025qwen3} & 8B & / & 78.8,~73.0&70.5,~52.1 &94.6,~92.3 & 75.7,~74.3 & 74.4,~73.5 & N/A & N/A & N/A & N/A & N/A & 19.3,~61.9 \\ 
Qwen3VL-thinking-A3B~\cite{team2025qwen3} & 30B & / & 82.6,~78.7 & 	69.5,~55.3 & 96.4,~94.9 &85.5,~84.5 & 79.0,~80.2 & N/A & N/A & N/A & N/A & N/A & 16.9,~58.9 \\ 

% IAD-R1~\textcolor{gray}{\tiny [arxiv 2025]} & 0.5B & 81.0 & 69.4 & & \underline{93.3} & \underline{95.5} & 88.6 & \underline{83.8} \\
% Anomaly-OV~\textcolor{gray}{\tiny [CVPR 2025]} & 0.5B & 50.0 & 50.0 &  & 50.0 & 53.8 & 50.0 & 50.6 \\
% Qwen2.5-VL-Instruct & 3B & 62.6 & 52.9 &  & 54.2 & 64.4 & 50.3 & 57.1 \\
% IAD-R1~\textcolor{gray}{\tiny [arxiv 2025]} & 3B & 77.6 & 59.2 & & 85.2 & 89.1 & 83.4 & 77.4 \\
% AnomalyR1~\textcolor{gray}{\tiny [arxiv 2025]} & 3B & 69.4 & 56.0 &  & 56.7 & 61.0 & 57.6 & 60.1 \\
InternVL-2.5~\cite{chen2024expanding} & 8B & / & 69.8,~72.9 & 63.7,~96.3 &81.0,~72.8 & 65.9,~63.1 & 68.7,~59.3 & N/A & N/A & N/A & N/A & N/A &	19.7,~57.5 \\
%  Anomaly-OV~\textcolor{gray}{\tiny [CVPR 2025]} & 7B & 74.3 & \underline{70.3} &  & 77.5 & 90.7 & \underline{88.7} & 78.9 \\
% InternVL-3~\cite{zhu2025internvl3}  & 9B & /  & 83.3,~84.4 & \textbf{89.5},~96.8 	&93.7,~91.0 &77.4,~75.5 & 72.4,~74.1 & N/A & N/A & N/A & N/A & N/A & 23.2,~65.2 \\ 
InternVL-3~\cite{zhu2025internvl3} & 78B & / &81.0,~83.4 &82.5,~88.6 & 91.4,~92.0 &75.2,~75.9 &74.7,~77.2 & N/A & N/A & N/A & N/A & N/A & 22.2,~70.1\\
MiniCPM-V-2.6~\cite{yao2024minicpm} & 8B & / & 70.3,~70.6 &	70.2,~97.1 & 76.8,~66.3 & 67.0,~64.2 & 67.0,~54.8 & N/A & N/A & N/A & N/A & N/A & 21.7,~68.3 \\
\midrule
% WinCLIP~\textcolor{gray}{\tiny [CVPR 2023]} & / & I+P   &  &  &  &  & &  & & & N/A \\
% FADE~\textcolor{gray}{\tiny [BMVC 2024]} & / & I+P  &  &  &  &  & &  & & & N/A \\
% AnomalyCLIP~\textcolor{gray}{\tiny [ICLR 2024]} & / & I+P  &  &  &  &  & &  & & & N/A \\
% AdaCLIP~\textcolor{gray}{\tiny [ECCV 2024]} & / & I+P  &  &  &  &  & &  & & & N/A \\
% AA-CLIP~\textcolor{gray}{\tiny [CVPR 2025]} & / & I+P  &  &  &  &  & &  & & & N/A \\

Triad~\cite{li2025triad} & 7B & T+I  &\underline{85.5},~83.8 & 79.8,~68.1 & 95.2,~95.4	& 82.5,~82.4 & 84.7,~89.4 & N/A & N/A & N/A & N/A & N/A & 8.6,~35.9
  \\
IAD-R1(Qwen2.5-VL)~\cite{li2025iad} & 3B &  T+I & /,~/  &/,83.4  & /,~89.1& /,~/  &/,~77.6 & N/A & N/A & N/A & N/A & N/A & /,~/ \\
IAD-R1(Qwen2.5-VL)~\cite{li2025iad} & 7B &  T+I & /,~/ &/,~85.4 & /,~90.8& /,~/  &/,~81.9 & N/A & N/A & N/A & N/A & N/A  & /,~/ \\

AnomalyGPT~\cite{gu2024anomalygpt} & 7B & T+I+P &71.1,~53.9 &72.7,~13.0 &75.7,~74.2 & 71.0,~56.2& 65.1,~72.2 &77.8,~\textbf{98.4}&	75.1,~99.9& 83.5,~98.9&	72.4,~98.6&	80.2,~96.4&11.9,~36.7 \\
% Myriad~\cite{li2023myriad} & 7B&T+I+P&&&&&86.2,~72.2&&&&&87.6, /\\
EIAD~\cite{zhang2025eiad} & 7B & T+I &68.4,~73.3&52.7,~87.2 &	83.4,~74.8 &59.6,~56.1 &  77.7,~75.2 &57.6,~91.4&52.1,~99.7 &61.2,~85.9 & 53.7,~96.7 &63.4,~83.5& 21.1,~65.7\\
% & GPT-4.1 & / & 81.9 & 66.7 & 69.1 & 81.8 & 90.1 & 79.9 & 78.3 \\
% & Claude-Sonnet-4 & / & 67.6 & 65.9 & 63.5 & 69.2 & 88.4 & 81.7 & 72.7 \\
\midrule
Qwen2.5-VL*~\cite{bai2025qwen2} & 3B & / & 52.9,~56.6 & 46.3,~69.9 & 	64.1,~51.3 & 50.0,~52.3 & 51.0,~52.8 & 55.6,~57.2 & 45.4,~10.6 & 71.4~91.8& 52.3,~79.9 & 53.4,~46.5 & 16.8,~49.0 \\

Qwen2.5-VL+R1*~\cite{guo2025deepseek} & 3B & I & 58.1,~60.5 &54.1,~94.8&57.9,~38.0 & 55.0,~51.2& 65.3,~57.8 & 72.6,~88.0 & 74.4,~99.9 & 85.6,~98.5 & 63.2,~97.0 & 67.1,~56.7 & 24.2,~71.7 \\

% Qwen2.5-VL~\cite{bai2025qwen2} & 3B & / & 41.6,~56.6 & 46.3,~69.9 & 	64.1,~51.3 & 5.0,~52.3 & 51.0,~52.8 & N/A & 45.4,10.6 & N/A & N/A & N/A & 16.8,~49.0 \\ 

Qwen2.5-VL+CGRO* & 3B & I & 67.6,~72.4  & 67.9,~82.5 & 78.3,~74.4 & 61.4,~59.9&62.7,~72.6 & 73.5,~94.3 & 75.2,~99.8 & 86.1,~98.4 & 62.1,~97.5 &70.6,~81.5& 25.6,~72.4\\
Qwen2.5-VL*~\cite{bai2025qwen2} & 7B & / &63.4,~63.3&65.0,~93.8&	68.3,~56.2 &60.5,~57.2 & 59.6,~45.8 & 61.7,~85.6 & 56.6,~99.8 & 76.0,~98.5 & 56.5,~70.3 & 57.5,~73.8 & 20.8,~65.5 \\ 
Qwen2.5-VL+R1*~\cite{guo2025deepseek} & 7B & I & 80.0,~82.0&	77.3,~97.2&90.7,~87.2&74.9,~74.2&77.1,~69.3 & \underline{78.5},~96.7 & 80.8,~99.8 & 92.7,~98.6 & 67.0,~97.6 & 73.6,~90.9 &  \underline{26.3},~\underline{73.8}\\

Qwen2.5-VL+CGRO* & 7B & I &83.9,~\underline{86.9}&81.4,~97.0&94.5,~93.6&79.9,~79.8&79.8,~77.3& \textbf{80.7},~\underline{97.1}&82.3,~99.8&94.2,~98.6&70.5,~97.8&75.6,~92.2&  \textbf{27.1},~\textbf{74.7}\\

% XXX(Qwen2.5-VL-Instruct) & 7B \\
\bottomrule
\end{tabular}%
\label{model_comparison-1}
\end{table*}
\section{Experiments}
\subsection{Experimental Settings}
For training, we curate a dataset of 4K industrial images collected from multiple public anomaly detection benchmarks, including VisA~\cite{zou2022spot}, GoodsAD~\cite{zhang2024pku}, Vision~\cite{bai2023vision}, PR-REAL~\cite{qin2023image}, Real-IAD~\cite{wang2024real}, and BSData~\cite{schlagenhauf2021industrial}. Each image is annotated only at the image level~(normal/anomaly), providing weak supervision for reinforcement learning fine-tuning under the proposed consistency-guided optimization. 

\noindent\textbf{Evaluation details.} To comprehensively evaluate the performance of our approach, we select four representative datasets: MVTec-AD~\cite{bergmann2019mvtec}, WFDD~\cite{chen2024unified}, SDD~\cite{tabernik2020segmentation} and DTD~\cite{aota2023zero}. These datasets simulate complex real-world industrial production scenarios, encompassing diverse materials, anomaly categories and operational environments.
We adopt the zero-shot industrial anomaly detection setting, where the model is trained exclusively on the auxiliary dataset without any overlap with the target test domains. This setup rigorously tests the generalization ability of the model to unseen anomaly categories and domains.

For quantitative evaluation, we employ standard metrics including AUROC, AUPR and ACC, which jointly assess the model’s capability to distinguish between normal and anomalous instances at both global and local levels. In addition, to measure the quality and interpretability of the generated reasoning, we construct a COT-evaluation dataset based on MVTec-AD, following the VisionR1~\cite{huang2025vision} paradigm. Each test image is paired with its corresponding ground-truth anomaly description synthesized from official annotations and expert-written explanations, forming the MVTec-COT benchmark. The generated reasoning is then evaluated using SBERT and ROUGE-L scores, which respectively capture semantic alignment with reference explanations and textual fidelity in terms of content overlap and recall. 
% Together, these metrics provide a comprehensive evaluation of both anomaly detection accuracy and the faithfulness of reasoning, enabling systematic analysis of the model’s capability to achieve interpretable, end-to-end anomaly understanding under image-level supervision.
Together, these metrics comprehensively evaluate detection accuracy and reasoning faithfulness, enabling systematic assessment of interpretable anomaly understanding under image-level supervision.

\noindent\textbf{Implementation details.} Our framework is implemented on top of Qwen2.5-VL-7B~\cite{bai2025qwen2}, a state-of-the-art open-source multimodal large language model that achieves an effective balance between reasoning capability, visual understanding, and computational efficiency. To ensure uniform processing across datasets, all images are resized to 420$\times$420 pixels during both training and inference. During fine-tuning, we adopt LoRA adaptation on the language and cross-modal layers while keeping the vision encoder frozen to preserve the pretrained visual representations. Each training batch consists of 16 samples, and for each input, 8 candidate generations are sampled per iteration to facilitate group-based reinforcement optimization. All experiments are conducted on two NVIDIA A6000 GPUs. 

\noindent\textbf{Comparison baselines.} We compare three families of approaches: commercial MLLMs, open-source MLLMs, and specialized anomaly detection frameworks. Commercial and open-source MLLMs include GPT-4o~\cite{hurst2024gpt}, GPT-4.1, GPT-5-nano, Gemini-2.5-Flash~\cite{comanici2025gemini}, InternVL-2.5~\cite{chen2024expanding} and InternVL-3~\cite{zhu2025internvl3}, MiniCPM-V-2.6~\cite{yao2024minicpm}, LLaVA-1.6~\cite{liu2024llava}, and LLaVA-OneVision~\cite{li2024llava}, as well as Qwen-2.5/3 VL~\cite{bai2025qwen2,team2025qwen3} variants. The specialized frameworks are divided by optimization strategy: SFT-optimized methods include AnomalyGPT~\cite{gu2024anomalygpt}, EIAD~\cite{zhang2025eiad}, and Triad~\cite{li2025triad}, while reinforcement-optimized methods include IAD-R1~\cite{li2025iad} and Ours. 
For consistency, we employ a unified prompting protocol with three tiers based on model capabilities: (1) models that reliably support structured outputs are asked to produce a reasoning process with predefined tags followed by a “Yes/No” answer; (2) models that cannot consistently follow tagged formats but can generate coherent reasoning are asked to produce a reasoning process followed by a “Yes/No” answer; (3) models with limited reasoning capability are asked to state whether an anomaly exists and briefly justify it.

\noindent\textbf{Reproduction protocol.} Triad~\cite{li2025triad}, AnomalyGPT~\cite{gu2024anomalygpt}, and EIAD~\cite{zhang2025eiad} are reproduced from their official codebases and evaluated on our benchmark. For EIAD~\cite{zhang2025eiad}, the model is instructed to output the reasoning and the final answer together with a predicted anomaly bounding box; this box is passed to a pretrained SAM~\cite{kirillov2023segment} to produce a segmentation mask, which serves as the anomaly map for pixel-level metrics. IAD-R1~\cite{li2025iad} is not open-sourced at the time of our experiments, so we report its results directly from the original paper, matching datasets and metrics where available.
\subsection{Experiment Performance}
Tab.~\ref{model_comparison-1} presents a comprehensive comparison across image-level detection, pixel-level localization, and reasoning quality. 
Trained exclusively with image-level labels via reinforcement learning, Our model achieves the highest reasoning scores among all competitors, demonstrating its strong ability to generate coherent, anomaly-aware explanations. 
For image-level detection, our model achieves strong performance, with ACC reaching the second-best average across all datasets, ranking only behind GPT-4.1 while using significantly less supervision. Most MLLMs report `N/A' for pixel-level metrics because they can output only text and cannot produce anomaly maps without external vision components. EIAD~\cite{zhang2025eiad} and AnomalyGPT~\cite{gu2024anomalygpt} circumvent this limitation by introducing additional modules, such as SAM~\cite{kirillov2023segment} in EIAD~\cite{zhang2025eiad} and a Vision Expert in AnomalyGPT, to obtain segmentation masks or anomaly heatmaps. These auxiliary models make localization possible but break end-to-end inference and add system complexity.
We apply ReAL to six model variants, spanning two model scales and three training settings:   
(i) \textit{Qwen2.5-VL-3B/7B}~\cite{bai2025qwen2}: the base model without any fine-tuning;  
(ii) \textit{Qwen2.5-VL-3B/7B+R1}~\cite{guo2025deepseek}: trained via reinforcement learning with format and accuracy rewards only;  
(iii) \textit{Qwen2.5-VL-3B/7B+CGRO}: further optimized with an additional consistency reward to align reasoning and visual attention, building upon the R1 objective.
ReAL enables all six models to output pixel-level anomaly maps without relying on any external segmentation or detection modules, achieving truly end-to-end anomaly localization. Across these variants, adding CGRO leads to consistent improvements in anomaly detection, localization, and reasoning. 

\begin{figure}[tp]
    \centering
    \includegraphics[width=0.95\linewidth]{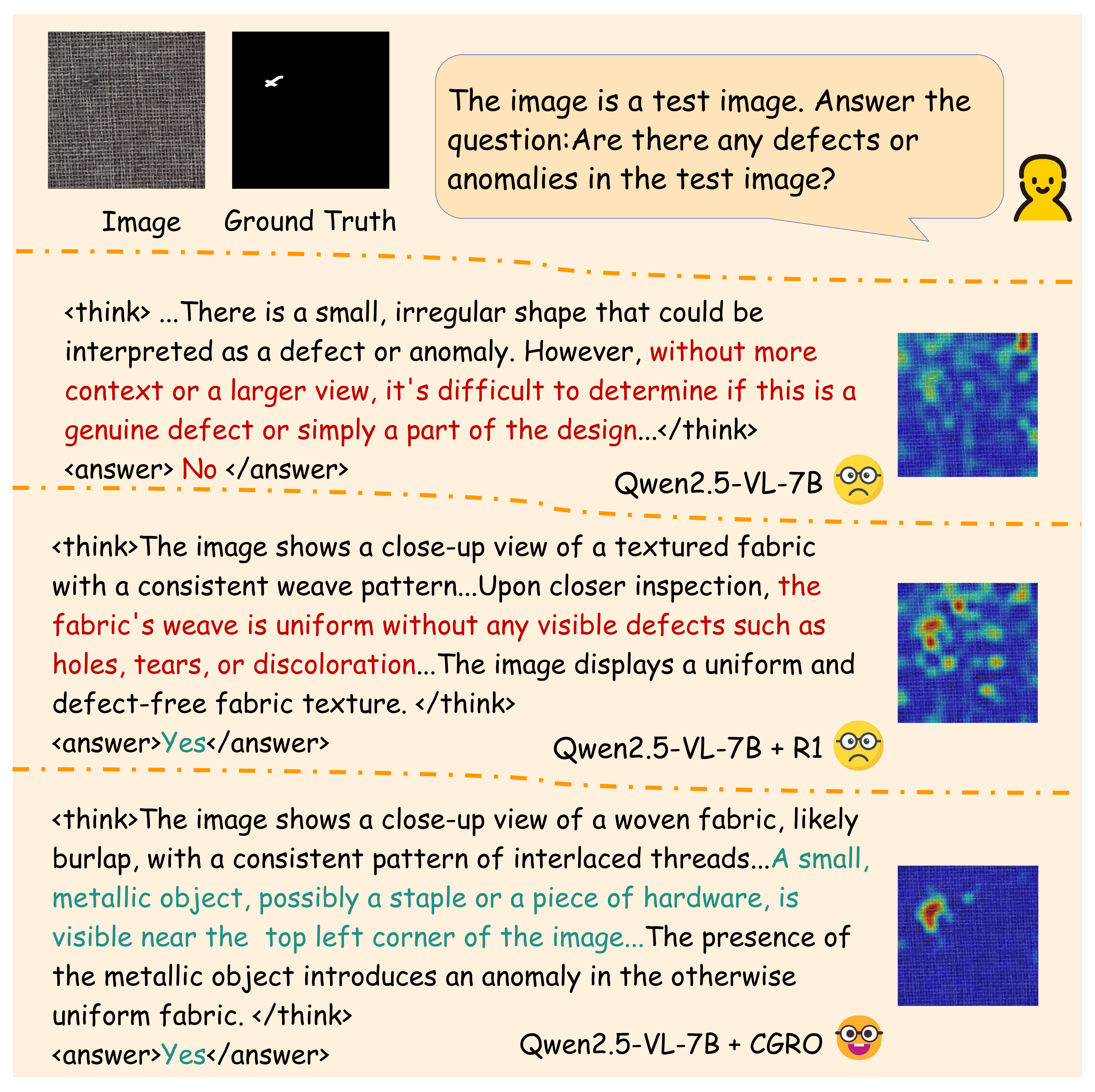}
    \caption{Qualitative comparison of \textit{Qwen2.5-VL-7B}, \textit{Qwen2.5-VL-7B+R1}, and \textit{Qwen2.5-VL-7B+CGRO}. } 
    \label{fig:consistency}
\end{figure}

\noindent\textbf{Qualitative comparisons.}
As Fig.~\ref{fig:consistency} shows, \textit{Qwen2.5-VL-7B} exhibits limited anomaly perception, frequently missing defects or producing irrelevant reasoning, resulting in poor detection and localization. Reinforcement learning substantially improves the model’s anomaly sensitivity in \textit{Qwen2.5-VL-7B+R1}, as evidenced by gains in both image-level accuracy and reasoning relevance. However, without explicit supervision on the reasoning process, the model often generates answers that contradict its rationale—for instance, predicting “anomaly present” while describing the image as visually normal. This inconsistency yields diffuse visual attention and degrades the quality of ReAL-based localization.
In contrast, \textit{Qwen2.5-VL-7B+CGRO} aligns reasoning with visual evidence, concentrating attention on true anomalies. The consistent gains in both detection and localization validate that the consistency reward effectively enforces coherent, visually grounded reasoning.

\begin{table}[!t]
\centering
\scriptsize
% \cAUPRtion{Ablation study on the impact of different components in anomaly detection. The model is based on the Qwen2.5-VL-7B. The evaluation is performed on both image-level and pixel-level detection tasks. }
\caption{Ablation on ReAL and CGRO. \textit{Vanilla} denotes the original \textit{Qwen2.5-VL-7B}. All metrics are macro-averaged across four AD datasets.}

\resizebox{\linewidth}{!}{
\begin{tabular}{cccccccccccccc}
\toprule
\multirow{2}{*}{\textbf{Model}}  & \multicolumn{3}{c}{\textbf{Image-level}} & \multicolumn{3}{c}{\textbf{Pixel-level}} \\
\cmidrule(lr){2-4} \cmidrule(lr){5-7} 
& \textbf{AUROC} & \textbf{AUPR} &\textbf{ACC} & \textbf{AUROC} & \textbf{AUPR} &\textbf{ACC} \\
\midrule
Vanilla & 63.4 & 59.4&63.3&64.7&2.6&73.0\\
Vanilla + ReAL & 63.4& 59.4& 63.3&61.7 & 5.1&85.6 \\
Vanilla + CGRO &  \textbf{83.9} & \textbf{76.7}&\textbf{86.9}&\underline{72.7}&\underline{5.9}&\underline{92.6}\\
Full &   \textbf{83.9} & \textbf{76.7}&\textbf{86.9}&\textbf{80.7}&\textbf{13.3}&\textbf{97.1} \\
\bottomrule
\end{tabular}%
}

\label{ablation_study_1}
\end{table}

\subsection{Ablation Study}
% \textbf{Component effectivity.}
\noindent\textbf{Ablation on ReAL and CGRO.}
We conduct an ablation study to evaluate the effectiveness of the two core modules in our framework: the Reasoning-Driven Anomaly Localization (ReAL) and the Consistency-Guided Reasoning Optimization (CGRO). The base model is \textit{Qwen2.5-VL-7B}~\cite{bai2025qwen2}, and all metrics in Tab.~\ref{ablation_study_1} are averaged across MVTec-AD~\cite{bergmann2019mvtec}, WFDD~\cite{chen2024unified}, SDD~\cite{tabernik2020segmentation} and DTD~\cite{aota2023zero}. Detailed results for each dataset are provided in the Supplementary Material.
Without ReAL, we average the visual attention of all reasoning tokens to form the anomaly map. This naive aggregation blurs spatial focus and weakens localization, reducing pixel-level AUROC from 80.7 to 72.7 and ACC from 97.1 to 92.6. With ReAL, the model selects anomaly-relevant reasoning tokens and aggregates their attention selectively, yielding a sharper and more accurate localization map. CGRO further enhances both detection and localization. Image-level AUROC rises from 63.4 to 83.9, and pixel-level AUROC increases from 61.7 to 80.7. Together, ReAL and CGRO lead to clear and consistent gains in all evaluation metrics, confirming the complementary roles of reasoning-driven localization and reasoning-visual alignment in our framework. 

% \begin{table}[htbp]
% \centering
% \scriptsize
% \caption{Ablation on token selection ReAL. Base model is \textit{Qwen2.5-VL-7B + CGRO}. We compare different ways of filtering and aggregating reasoning tokens for pixel-level localization. $S_\text{T}$ denotes the semantic relevance score measuring alignment with anomaly concepts, and $S_\text{I}$ denotes the spatial entropy score reflecting attention concentration. Results are pixel-level metrics averaged across four datasets.}
% \begin{tabular}{ccccc}
% \toprule
% Spatial entropy \textbf{$S_\text{I}$} & Semantic relevant \textbf{$S_\text{T}$} & \textbf{AUROC} & \textbf{ACC} \\
% \midrule
%  &  & 72.7 & 74.6 \\
% \checkmark &  & 57.7 & 89.2 \\
%  &\checkmark  & \underline{76.7} & \underline{90.2} \\
% \checkmark & \checkmark & \textbf{80.7} & \textbf{97.1} \\
% \bottomrule
% \end{tabular}
% \label{ablation_rdam}
% \end{table}

\begin{table}[htbp]
\centering
\scriptsize
\caption{Ablation on token selection in ReAL. Base model is \textit{Qwen2.5-VL-7B + CGRO}. We compare different ways of filtering and aggregating reasoning tokens for pixel-level localization. $S_\text{T}$ denotes the semantic relevance score measuring alignment with anomaly concepts, and $S_\text{I}$ denotes the spatial entropy score reflecting attention concentration. Results are pixel-level metrics averaged across four datasets.}
\begin{tabular}{cccccc}
\toprule

Spatial entropy \textbf{$S_\text{I}$} & Semantic relevant \textbf{$S_\text{T}$} & \textbf{AUROC} & \textbf{AUPR}&\textbf{ACC} \\
% \textbf{$S_\text{I}$} & Semantic relevant \textbf{$S_\text{T}$} & \textbf{AUROC} & \textbf{AUPR}&\textbf{ACC} \\

\midrule
 &  & 72.7 & 5.9&\underline{92.6} \\
\checkmark &  & 74.1 & 11.8&89.2 \\
 &\checkmark  & \underline{76.7} & \textbf{14.0}&90.2 \\
\checkmark & \checkmark & \textbf{80.7} & \underline{13.3}&\textbf{97.1} \\
\bottomrule
\end{tabular}
\label{ablation_rdam}
\end{table}

\begin{figure}[tp]
    \centering
    \includegraphics[width=0.95\linewidth]{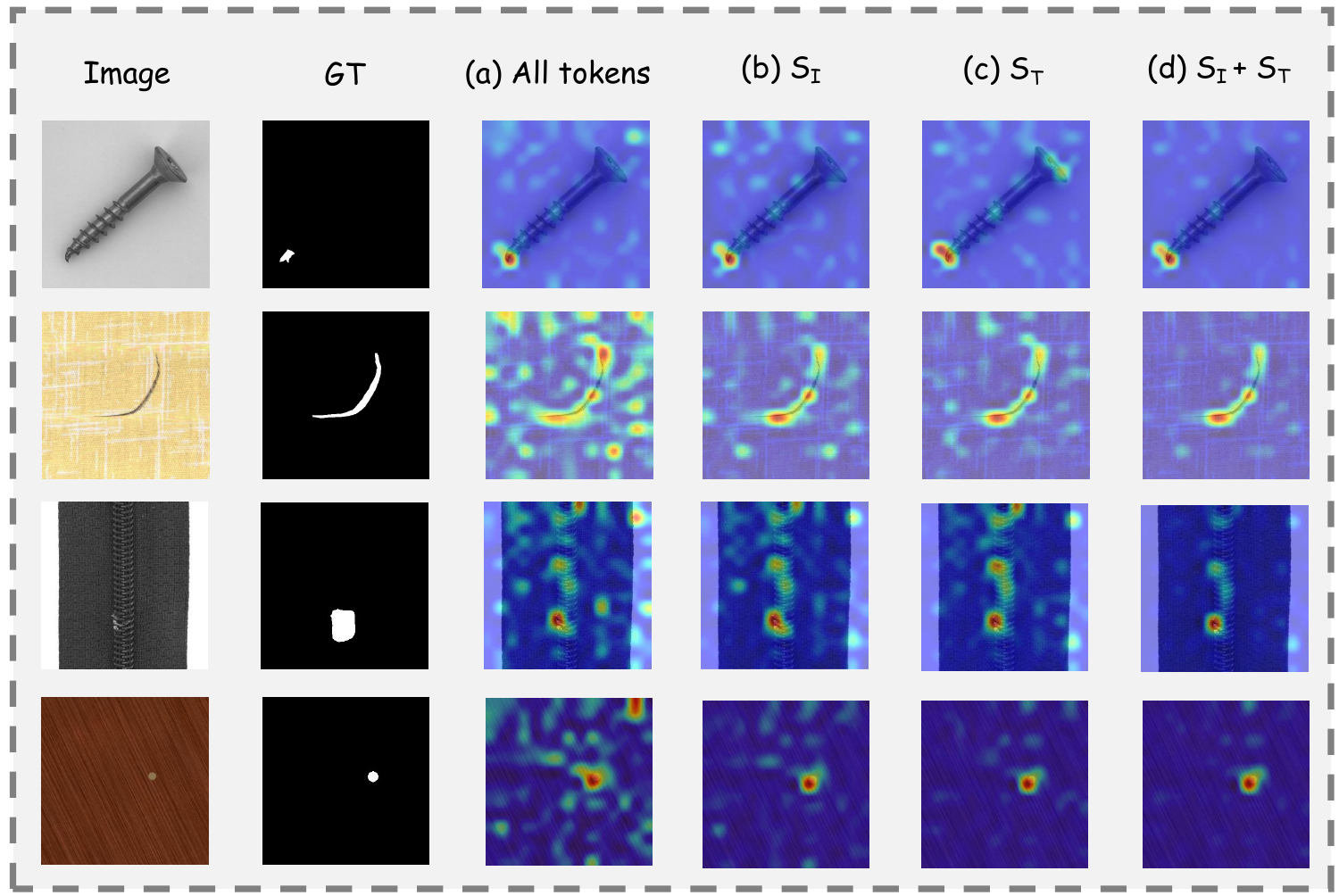}
    \caption{Visualization of different token selection used in ReAL. Each column shows anomaly maps generated by: (a) keeping all tokens  (b) filtering with spatial entropy $S_\text{I}$, (c) filtering with semantic relevance $S_\text{T}$,  (d) combining both scores with weighted aggregation. $S_\text{T}$ measures semantic alignment with anomaly-related concepts, while $S_\text{I}$ measures attention concentration. Methods that incorporate both criteria yield more accurate anomaly maps.} 
    \label{fig:aggregate}
\end{figure}

% \noindent\textbf{Effectiveness of ReAL.}
\noindent\textbf{Ablation on token selection in ReAL.}
Tab.~\ref{ablation_rdam} presents an ablation study of our Reasoning-Driven Anomaly Localization pipeline. 
We report the average pixel-level localization performance over the four datasets. 
Using all tokens without any filtering yields relatively low AUROC and ACC. 
% When we add POS-based filtering, we remove many irrelevant tokens and observe clear gains in AUPR and ACC. 
When we apply $S_\text{I}$ filtering with spatial entropy, many spatially irrelevant tokens are removed and we observe clear gains in ACC.
Applying $S_\text{T}$ filtering with
semantic relevance further improves AUROC and ACC, showing the benefit of retaining tokens aligned with anomaly semantics.
Finally, adding both $S_\text{I}$ and $S_\text{T}$ using composite weights enhances overall performance, achieving an AUROC of 80.7 and an ACC of 97.1, which confirms that the two criteria contribute complementary spatial and semantic cues for selecting anomaly-relevant reasoning tokens.
As illustrated in Fig.~\ref{fig:aggregate}, this configuration produces cleaner and more focused anomaly maps, which supports the effectiveness of our token selection and weighting strategy.

\section{Conclusion}
We introduced a reasoning-driven framework that achieves end-to-end anomaly detection and localization using only image-level supervision. By extracting anomaly-relevant reasoning signals with ReAL and aligning the reasoning process through consistency-guided optimization, the model learns to produce coherent anomaly descriptions and accurate localization without pixel-level labels or external modules. Experiments across multiple benchmarks show strong generalization and competitive performance against models trained with far richer supervision. These results highlight the potential of reasoning-guided multimodal learning for scalable and interpretable industrial anomaly understanding.
In the future, we will extend the evaluation to a broader range of MLLMs to further validate the generality and robustness of the proposed framework.
\section*{Acknowledgements}
This work is supported by Beijing Municipal Science and Technology Commission, Administrative Commission of Zhongguancun Science Park~(L242105).

{
    \small
    \bibliographystyle{ieeenat_fullname}
    \bibliography{main}
}
\clearpage
\setcounter{page}{1}
\setcounter{section}{6} % 将当前章节编号重置为0
\setcounter{table}{3}
\setcounter{figure}{4}
% \section{新的章节}

% \appendix
% \renewcommand\thetable{\Alph{table}}
% \renewcommand\thefigure{\Alph{figure}}
\maketitlesupplementary

\begin{figure*}[tp]
    \centering
    \includegraphics[width=0.95\linewidth]{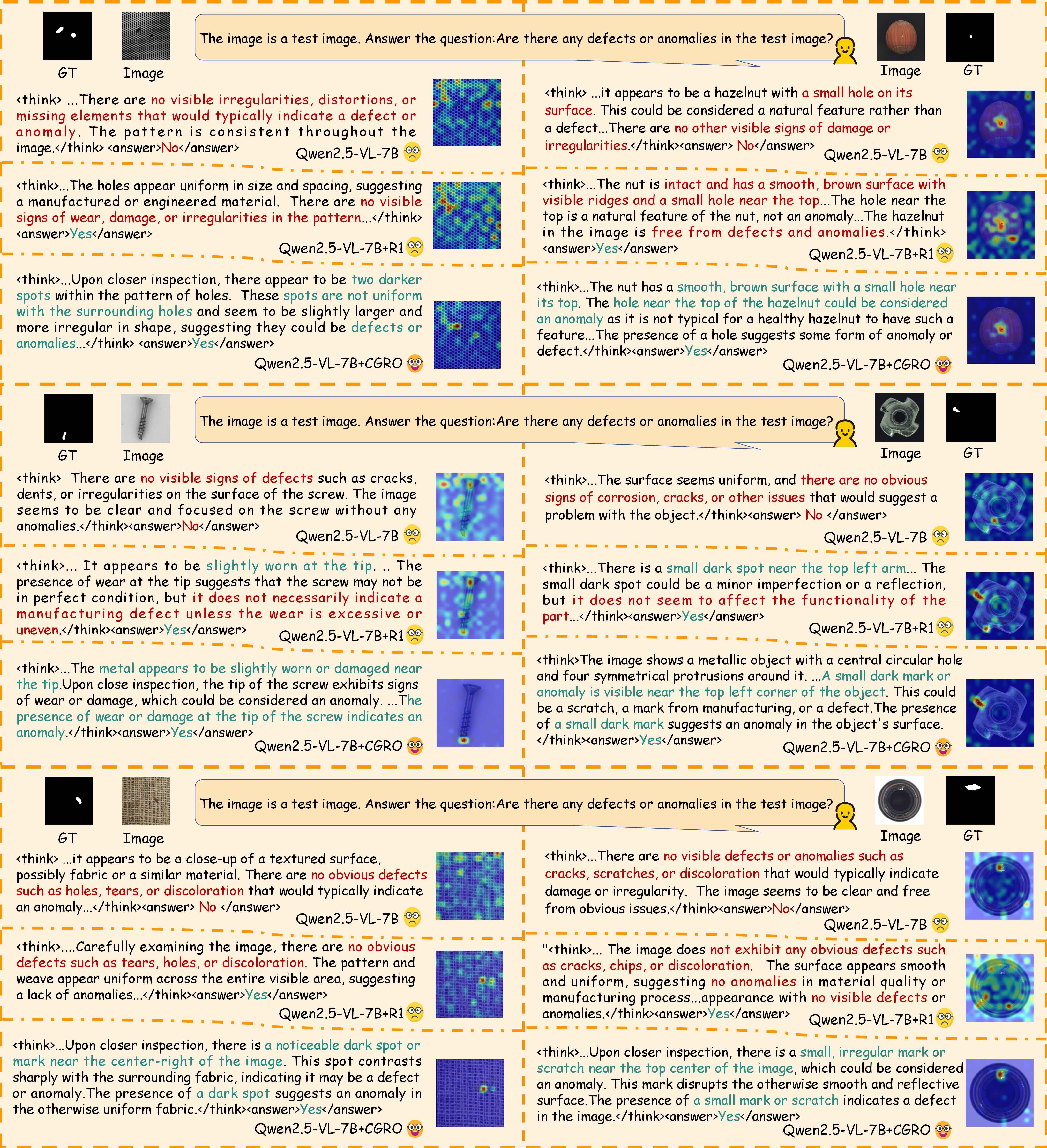}
    \caption{Qualitative comparison of \textit{Qwen2.5-VL-7B}, \textit{Qwen2.5-VL-7B+R1}, and \textit{Qwen2.5-VL-7B+CGRO}.} 
    \label{fig:consistency-ablation}
\end{figure*}
\begin{figure*}[tp]
    \centering
    \includegraphics[width=0.95\linewidth]{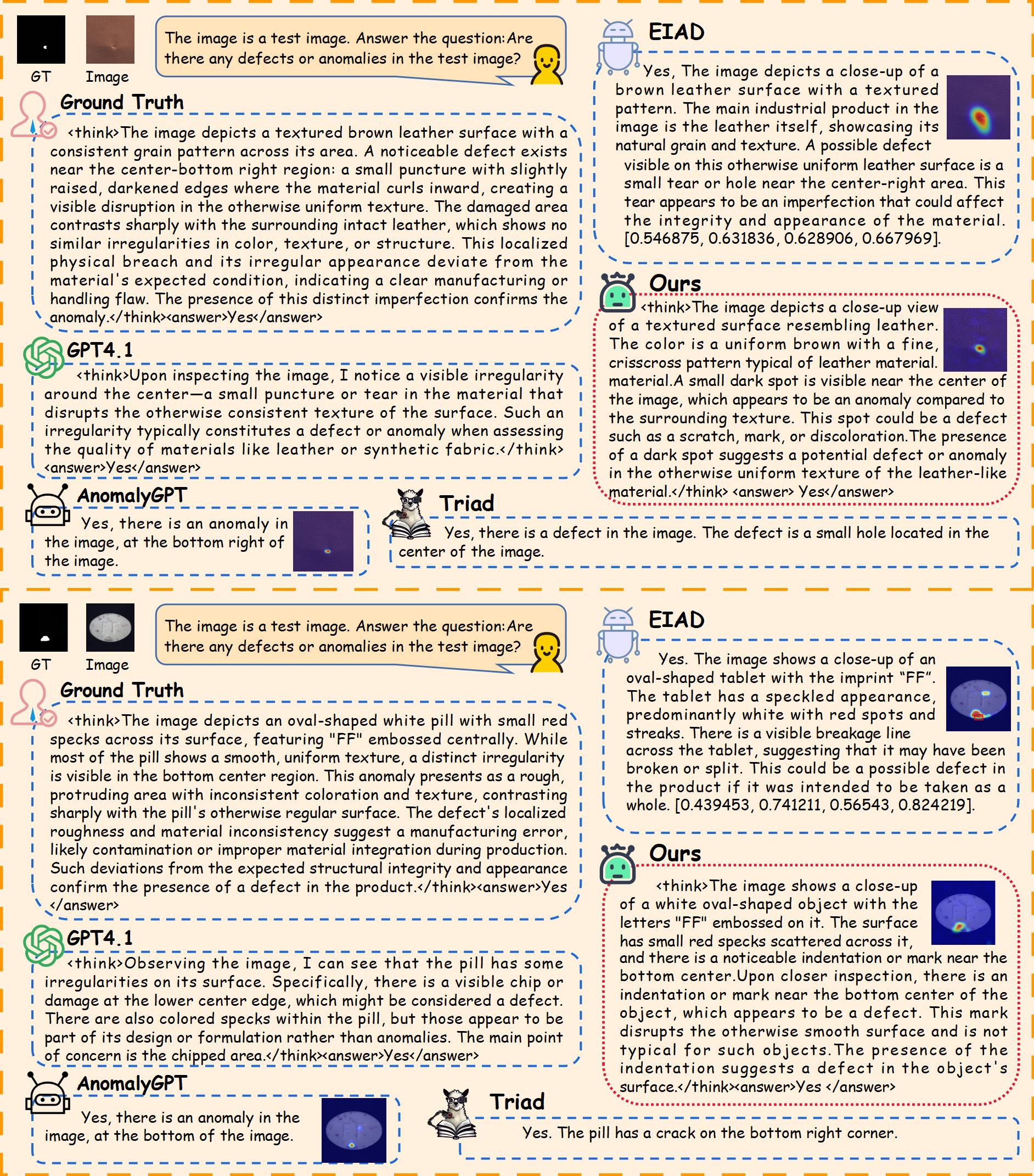}
    \caption{Comparison of GPT-4.1, Triad, AnomalyGPT, EIAD, and our method.} 
    \label{fig:comparison-with-others}
\end{figure*}
% \begin{table}[!t]
% \centering
% \scriptsize
% % \cAUPRtion{Ablation study on the impact of different components in anomaly detection. The model is based on the Qwen2.5-VL-7B. The evaluation is performed on both image-level and pixel-level detection tasks. }
% \caption{Ablation study of the CGRO hyperparameters, including the number of top-$k$ reasoning tokens, the Jaccard threshold $\delta_1$ and $\delta_2$. Base model is \textit{Qwen2.5-VL-7B + CGRO}. Results are image-level and pixel-level metrics on the MFDD dataset.}

% \resizebox{\linewidth}{!}{
% \begin{tabular}{cccccccccccccc}
% \toprule
% \multirow{2}{*}{\textbf{Model}}  & \multicolumn{3}{c}{\textbf{Image-level}} & \multicolumn{3}{c}{\textbf{Pixel-level}} \\
% \cmidrule(lr){2-4} \cmidrule(lr){5-7} 
% & \textbf{AUROC} & \textbf{AUPR} &\textbf{ACC} & \textbf{AUROC} & \textbf{AUPR} &\textbf{ACC} \\
% \midrule
% Vanilla & 63.4 & 59.4&63.3&64.7&2.6&73.0\\
% Vanilla + ReAL & 63.4& 59.4& 63.3&61.7 & 5.1&\underline{85.6} \\
% Vanilla + CGRO &  \textbf{83.9} & \textbf{76.7}&\textbf{86.9}&\underline{72.7}&\underline{5.9}&74.6\\
% Full &   \textbf{83.9} & \textbf{76.7}&\textbf{86.9}&\textbf{81.0}&\textbf{13.3}&\textbf{97.1} \\
% \bottomrule
% \end{tabular}%
% }

% \label{CGRO:threshold}
% \end{table}

\begin{table}[!t]
\centering
\scriptsize
% \cAUPRtion{Ablation study on the impact of different components in anomaly detection. The model is based on the Qwen2.5-VL-7B. The evaluation is performed on both image-level and pixel-level detection tasks. }
\caption{Ablation study of the CGRO hyperparameters, including the number of top-$k$ reasoning tokens, the Jaccard threshold $\delta_1$ and $\delta_2$. Base model is \textit{Qwen2.5-VL-7B + CGRO}. Results are image-level and pixel-level metrics on the WFDD dataset.}

\resizebox{\linewidth}{!}{
\begin{tabular}{cccccccccccccc}
\toprule
\multicolumn{3}{c}{\textbf{hyperparameter}}  & \multicolumn{3}{c}{\textbf{Image-level}} & \multicolumn{3}{c}{\textbf{Pixel-level}} \\
\cmidrule(lr){1-3}\cmidrule(lr){4-6} \cmidrule(lr){7-9} 
$\delta_1$&$\delta_2$&$k$& \textbf{AUROC} & \textbf{AUPR} &\textbf{ACC} & \textbf{AUROC} & \textbf{AUPR} &\textbf{ACC} \\
\midrule
0.3&0.1&2&76.0&76.0&75.6&67.9&5.6&97.5\\
0.3&0.1&4&76.1&76.6&75.8&67.0&5.4&\underline{97.6}\\
0.3&0.3&3&\underline{77.8}&\underline{77.7}&\underline{77.1}&68.7&5.9&\underline{97.6}\\
0.5&0.1&3&73.3&74.4&72.1&\underline{69.5}&\underline{6.3}&97.5\\
0.3&0.1&3&\textbf{79.9}&\textbf{78.8}&\textbf{79.8}&\textbf{70.5}&\textbf{7.4}&\textbf{97.8}\\
% Vanilla + ReAL & 63.4& 59.4& 63.3&61.7 & 5.1&\underline{85.6} \\
% Vanilla + CGRO &  \textbf{83.9} & \textbf{76.7}&\textbf{86.9}&\underline{72.7}&\underline{5.9}&74.6\\
% Full &   \textbf{83.9} & \textbf{76.7}&\textbf{86.9}&\textbf{81.0}&\textbf{13.3}&\textbf{97.1} \\
\bottomrule
\end{tabular}%
}

\label{CGRO:threshold}
\end{table}

% \begin{table}[htbp]
% \centering
% \scriptsize
% \caption{Ablation on token selection ReAL. Base model is \textit{Qwen2.5-VL-7B + CGRO}. We compare different ways of filtering and aggregating reasoning tokens for pixel-level localization. $S_\text{T}$ denotes the semantic relevance score measuring alignment with anomaly concepts, and $S_\text{I}$ denotes the spatial entropy score reflecting attention concentration. Results are pixel-level metrics averaged across four datasets.}
% \begin{tabular}{ccccc}
% \toprule
% Spatial entropy \textbf{$S_\text{I}$} & Semantic relevant \textbf{$S_\text{T}$} & \textbf{AUROC} & \textbf{ACC} \\
% \midrule
%  &  & 72.7 & 74.6 \\
% \checkmark &  & 57.7 & 89.2 \\
%  &\checkmark  & \underline{76.7} & \underline{90.2} \\
% \checkmark & \checkmark & \textbf{81.0} & \textbf{97.1} \\
% \bottomrule
% \end{tabular}
% \label{ablation_rdam}
% \end{table}

\begin{table}[t]
\centering
\scriptsize
\caption{Ablation study comparing different threshold configurations for $\tau_\text{i}$ and $\tau_\text{t}$. Base model is \textit{Qwen2.5-VL-7B + CGRO}. Results are pixel-level metrics on the WFDD dataset.}
\begin{tabular}{cccccc}
\toprule

$\tau_\text{i}$ & $\tau_\text{t}$ & \textbf{AUROC} & \textbf{AUPR}&\textbf{ACC} \\
% \textbf{$S_\text{I}$} & Semantic relevant \textbf{$S_\text{T}$} & \textbf{AUROC} & \textbf{AUPR}&\textbf{ACC} \\

\midrule
Maximum Curvature & Maximum Curvature & \textbf{70.7} & 6.5&97.5 \\
Median & Maximum Curvature & 70.4 & 6.3&97.5 \\
Median & Median  & 70.4 & \textbf{7.5}&\underline{97.6} \\
Maximum Curvature & Median & \underline{70.5} & \underline{7.4}&\textbf{97.8} \\
\bottomrule
\end{tabular}
\label{ReAL:threshold}
\end{table}

% 增加Anomaly-R1 Anomaly-OV和IAD-R1 (Qwen3b)后的表格
\begin{table*}[t]
\centering
\tiny 
\caption{Dataset-wise results image-level and pixel-level performance for the ReAL and CGRO ablation study. Ablation on ReAL and CGRO. \textit{Vanilla} denotes the original \textit{Qwen2.5-VL-7B}. All metrics are macro-averaged across four AD datasets. Detection results are reported as (AUROC, AUPR).}

\resizebox{\linewidth}{!}{
\begin{tabular}{cccccccccccccc}
\toprule
\multirow{2}{*}{\textbf{Model}}  & \multicolumn{5}{c}{\textbf{Image-level}} & \multicolumn{5}{c}{\textbf{Pixel-level}} \\
\cmidrule(lr){2-6}  \cmidrule(lr){7-11} 
& \underline{\textbf{AVG}}&\textbf{SDD} & \textbf{DTD} & \textbf{WFDD} & \textbf{MVTec} & \underline{\textbf{AVG}} &\textbf{SDD} & \textbf{DTD} & \textbf{WFDD} & \textbf{MVTec} \\
\midrule
% Vanilla & 63.4 & 59.4&63.3&64.7&2.6&73.0\\
% Vanilla + ReAL & 63.4& 59.4& 63.3&61.7 & 5.1&\underline{85.6} \\
% Vanilla + CGRO &  \textbf{83.9} & \textbf{76.7}&\textbf{86.9}&\underline{72.7}&\underline{5.9}&74.6\\
% Full &   \textbf{83.9} & \textbf{76.7}&\textbf{86.9}&\textbf{80.7}&\textbf{13.3}&\textbf{97.1} \\
Vanilla & 63.4,~59.4&65.0,~13.7&68.3,~82.3& 60.5,~63.7& 59.6,~77.7& 64.7,~2.6& 60.6,~0.1&79.7,~4.7&58.3,~1.6&60.1,~4.1\\

Vanilla + ReAL & 63.4,~59.4&65.0,~13.7&68.3,~82.3&60.5,~63.7&59.6,~77.7& 61.7,~5.1& 56.6,~0.1&76.0,~13.2& 56.5,~2.0&57.5,~5.0 \\

Vanilla + CGRO & \textbf{83.9},~\textbf{76.7}&81.4,~44.3& 94.5,~96.4&79.9,~78.8&79.8,~87.1&\underline{72.7},~\underline{5.9}&  72.5,~0.2&87.1,~13.1& 63.3,~2.7&67.9,~7.5        \\

Full &   \textbf{83.9},~\textbf{76.7}  &  81.4,~44.3& 94.5,~96.4&79.9,~78.8&79.8,~87.1&\textbf{80.7},~\textbf{13.3}&  82.3,~2.6& 94.2,~26.4&70.5,~7.4&   75.6,~16.6     \\
\bottomrule
\end{tabular}%
}
\label{tab2s}
\end{table*}

% 增加Anomaly-R1 Anomaly-OV和IAD-R1 (Qwen3b)后的表格
\begin{table*}[t]
\centering
\tiny 
\caption{Dataset-wise pixel-level results for the ablation on token selection in the ReAL module. Base model is  \textit{Qwen2.5-VL-7B + CGRO}. We compare four configurations: using all tokens, filtering with spatial entropy $S_\text{I}$, filtering with semantic relevance $S_\text{T}$, and combining both criteria. Results are pixel-level metrics reported on MVTec-AD, WFDD, SDD, and DTD, complementing Table 3 in the main paper.}
\begin{tabular}{lcccccccccccccccccccc}
\toprule
\multirow{2}{*}{Spatial entropy \textbf{$S_\text{I}$}} & \multirow{2}{*}{Semantic relevant \textbf{$S_\text{T}$}} & \multicolumn{3}{c}{\underline{\textbf{AVG}}} &  \multicolumn{3}{c}{\textbf{SDD}}&\multicolumn{3}{c}{\textbf{DTD}}&\multicolumn{3}{c}{\textbf{WFDD}} &\multicolumn{3}{c}{\textbf{MVTec}}\\
\cmidrule(lr){3-5} \cmidrule(lr){6-8} \cmidrule(lr){9-11} \cmidrule(lr){12-14} \cmidrule(lr){15-17} 
&&\textbf{AUROC}&\textbf{AUPR}&\textbf{ACC}&\textbf{AUROC}&\textbf{AUPR}&\textbf{ACC}&\textbf{AUROC}&\textbf{AUPR}&\textbf{ACC}&\textbf{AUROC}&\textbf{AUPR}&\textbf{ACC}&\textbf{AUROC}&\textbf{AUPR}&\textbf{ACC}\\
 % \multirow{2}{*}{\textbf{Model}}  & \multicolumn{5}{c}{\textbf{Image-level}} & \multicolumn{5}{c}{\textbf{Pixel-level}}   \\

% & \underline{\textbf{AVG}}&\textbf{SDD} & \textbf{DTD} & \textbf{WFDD} & \textbf{MVTec} & \underline{\textbf{AVG}} &\textbf{SDD} & \textbf{DTD} & \textbf{WFDD} & \textbf{MVTec}  \\
\midrule
 &  & 72.7 & 5.9&\underline{92.6}&72.5&0.2&97.0&87.1&13.1&97.9&63.3&2.7&95.6&67.9&7.5&80.0 \\
 
\checkmark &  & 74.1 & 11.8&89.2&75.0&1.0&99.7&92.4&31.1&98.6 &55.0&1.7&68.6&74.0&13.4&90.0\\

 &\checkmark  & \underline{76.7} & \textbf{14.0}&90.2 &75.2&1.5&99.8&93.5&35.0&98.6&60.4&3.3&72.1&77.5&16.3&90.3\\
 
\checkmark & \checkmark & \textbf{80.7} & \underline{13.3}&\textbf{97.1} &82.3&2.6&99.8&94.2&26.4&98.6&70.5&7.4&97.8&75.6&16.6&92.2\\

% XXX(Qwen2.5-VL-Instruct) & 7B \\
\bottomrule
\end{tabular}%
\label{tab3s}
\end{table*}
\begin{figure}[t]
    \centering
    \includegraphics[width=0.99\linewidth]{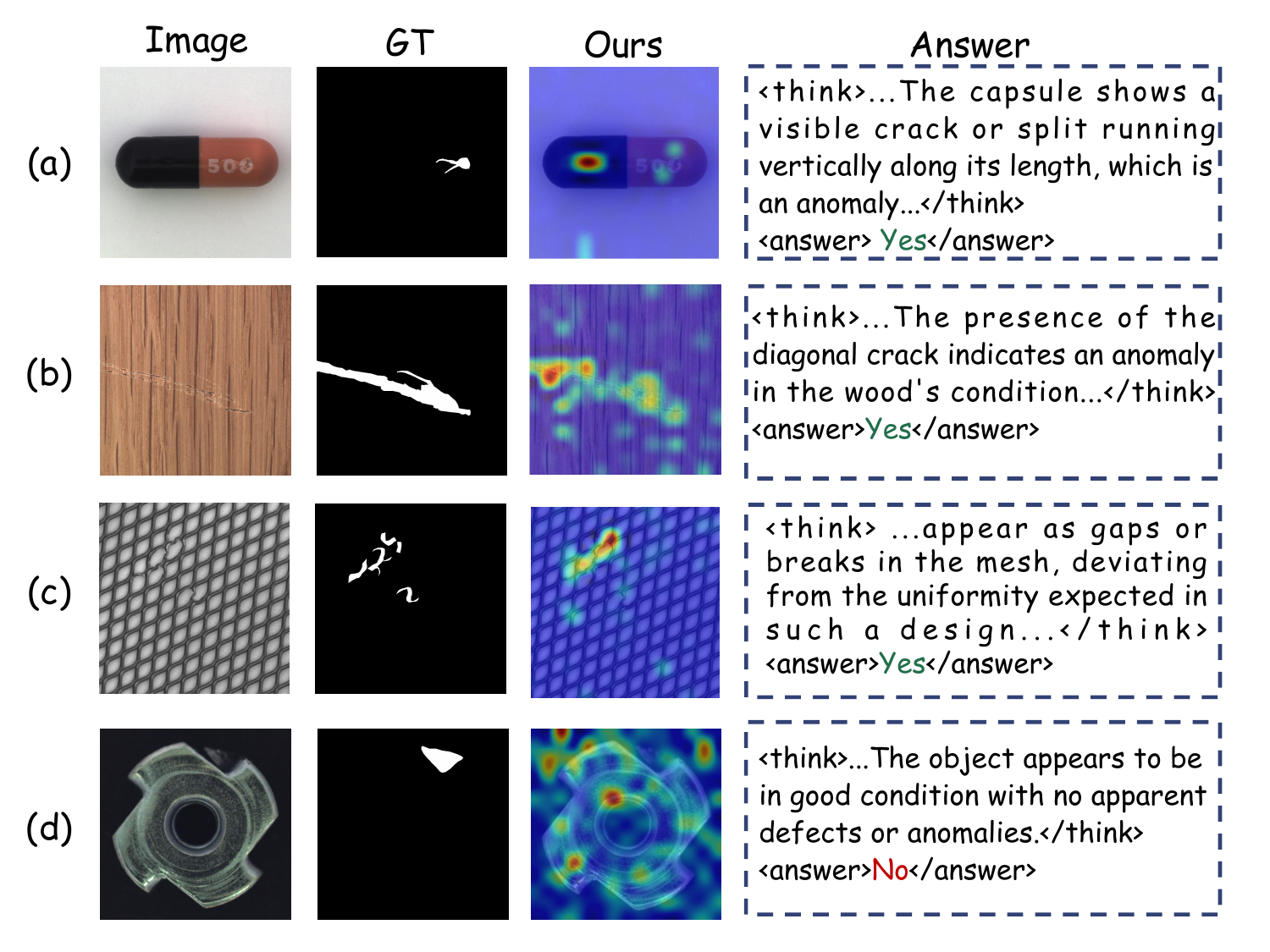}
    \caption{Typical failure cases. (a) Attention distracted by salient but non-defective regions; (b) coarser predicted anomaly boundaries; (c) missed detections in multi-spot scenarios; and (d) overall localization failure caused by incorrect image-level predictions and the subsequent lack of valid reasoning.} 
    \label{fig:failure-case}
\end{figure}

\section{Additional Qualitative Results}
\noindent\textbf{Impact of consistency reward.}
As shown in Fig.~\ref{fig:consistency-ablation}, additional qualitative results highlight the impact of the consistency reward. By introducing the consistency reward, the model aligns reasoning with visual evidence, focusing attention on true anomalies. This alignment leads to consistent improvements in both detection and localization, demonstrating that the consistency reward effectively enforces coherent, visually grounded reasoning, which enhances anomaly localization performance.

\noindent\textbf{Comparison with other methods.}
Fig.~\ref{fig:comparison-with-others} presents a qualitative comparison of GPT-4.1~\cite{hurst2024gpt}, Triad~\cite{li2025triad}, AnomalyGPT~\cite{gu2024anomalygpt}, EIAD~\cite{zhang2025eiad}, and our method. GPT-4.1~\cite{hurst2024gpt} and Triad~\cite{li2025triad} exhibit strong reasoning capabilities but fail to localize anomalies at the pixel level. In contrast, our method, trained solely with image-level supervision, achieves accurate anomaly detection and interpretable reasoning, while generating high-fidelity pixel-level localization maps. Notably, our method yields qualitative pixel-level localization on par with AnomalyGPT~\cite{hurst2024gpt} and EIAD~\cite{zhang2025eiad}, despite their reliance on additional vision modules and pixel-level supervision.
\section{Additional Ablation Studies}
\noindent\textbf{Hyperparameter ablation for CGRO.}
The CGRO module has three hyperparameters:  
(1) $k$, the number of top reasoning tokens used for spatial consensus,  
(2) $\delta_1$, a Jaccard threshold enforcing high consensus on anomalous images, and  
(3) $\delta_2$, a threshold enforcing low consensus on normal images.  
Larger $k$ increases token coverage but risks including noisy reasoning; $\delta_1$ and $\delta_2$ jointly shape the reward landscape to favor discriminative reasoning patterns.

Tab.~\ref{CGRO:threshold} reports performance over varying $(k, \delta_1, \delta_2)$. We select $k\!=\!3$, $\delta_1\!=\!0.3$, and $\delta_2\!=\!0.1$, which yield the most consistent gains across both image-level and pixel-level anomaly detection metrics.

\noindent\textbf{Hyperparameter ablation for ReAL.}
The ReAL module involves two thresholds: 
$\tau_\text{i}$, which filters tokens by spatial entropy (lower entropy $\Rightarrow$ more spatially concentrated attention), and  
$\tau_\text{t}$, which filters by semantic relevance (higher $S_\text{T}^r \Rightarrow$ stronger alignment with the anomaly query). 
To select these thresholds, we sort the tokens in ascending order of $S_\text{I}^r$ and $S_\text{T}^r$. We then determine the thresholds based on the maximum curvature and median strategies.

As shown in Tab.~\ref{ReAL:threshold}, we ultimately select $\tau_\text{i}$ as the threshold determined by maximum curvature and $\tau_\text{t}$ as  the median. 

\noindent\textbf{Ablation on ReAL and CGRO.}
To complement the averaged results reported in Table 2 of the main paper, we provide the complete dataset-wise performance metrics for the ablation study on ReAL and CGRO. As shown in Tab.~\ref{tab2s}, ReAL consistently improves localization by extracting anomaly-related tokens from the autoregressive reasoning process and aggregating their attention responses to produce pixel-level anomaly maps, which effectively enhances the model’s pixel-level segmentation capability. Building on this, CGRO leverages reinforcement learning to further strengthen both detection and localization by aligning the reasoning process with visual evidence. The dataset-level breakdown shows clear gains at both the image level and the pixel level, demonstrating the effectiveness of our method and the complementary roles of ReAL and CGRO within the overall framework.

\noindent\textbf{Ablation on token selection in ReAL.}
To complement the averaged results reported in Table 3 of the main paper, we provide the full dataset-wise pixel-level localization metrics for the ablation on token selection in the ReAL module. As shown in Tab.~\ref{tab3s}, combining both semantic relevance $S_\text{T}$ and spatial entropy $S_\text{I}$ with composite weighting leads to the strongest and most stable performance across all datasets, confirming that spatial concentration and semantic relevance provide complementary cues for selecting anomaly-relevant reasoning tokens.
\section{Discussion}
\noindent\textbf{MVTec-COT Construction.}
% \section{MVTec-COT Construction}
Following the established VisionR1 pipeline~\cite{huang2025vision}, our data generation process incorporates a rigorous human-verification stage to filter reasoning signals and minimize noise, thereby ensuring high-quality data synthesis. All anomaly-related concepts—including scene context, defect types, and spatial locations—are grounded in the publicly annotated MMAD benchmark~\cite{jiang2025mmad} to guaranty biological and physical validity. To ensure the integrity of our results, the MVTec-COT dataset is strictly reserved for evaluation and is never utilized during training or optimization, effectively preventing data leakage. Furthermore, the complete dataset will be publicly released to support reproducibility and further research in the community.

\noindent \textbf{Failure cases.}
Although our framework achieves competitive performance using only image-level supervision, we observe several typical failure cases (Fig.~\ref{fig:failure-case}). These include the model's attention being distracted by salient but non-defective regions, coarser predicted anomaly boundaries compared to dense-supervision methods, and missed detections in multi-spot scenarios. Furthermore, incorrect image-level predictions inherently prevent the generation of a valid reasoning process, ultimately leading to an overall localization failure.

\noindent \textbf{Limitations.}
Our framework reflects a trade-off between annotation efficiency and localization performance. While relying solely on image-level supervision and avoiding auxiliary vision modules improves scalability, a performance gap remains compared with fully supervised methods, particularly in boundary fidelity and attention misalignment in complex scenes. In addition, real-time deployment is constrained by the autoregressive decoding latency of MLLM backbones, as reasoning-based inference remains slower than conventional industrial anomaly detection pipelines.

%     \small
%     \bibliographystyle{ieeenat_fullname}
%     \bibliography{main}
% }


\end{document}